\documentclass[sigconf]{acmart}
\usepackage{mdframed}
\AtBeginDocument{%
  }

\setcopyright{acmlicensed}
\copyrightyear{2024}
\acmYear{2024}
\acmDOI{XXXXXXX.XXXXXXX}

\acmConference[XXX]{XXX}{XXX}{XXX}
\acmISBN{978-1-4503-XXXX-X/18/06}

\begin{document}

\title[Real-Time Adaptive Industrial Robots]{Real-Time Adaptive Industrial Robots: Improving Safety And Comfort In Human-Robot Collaboration}


\author{Damian Hostettler}

\email{damian.hostettler@student.unisg.ch}
\affiliation{%
  \institution{Institute of Computer Science, University of St. Gallen}
  \city{St. Gallen}
  \country{Switzerland}
}

\author{Simon Mayer}
\affiliation{%
  \institution{Institute of Computer Science, University of St. Gallen}
  \city{St. Gallen}
  \country{Switzerland}
\email{simon.mayer@unisg.ch}
}

\author{Jan Liam Albert}
\affiliation{%
  \institution{Institute of Computer Science, University of St. Gallen}
  \city{St. Gallen}
  \country{Switzerland}
\email{janliam.albert@unisg.ch}
}

\author{Kay Erik Jenss}
\affiliation{%
  \institution{Institute of Computer Science, University of St. Gallen}
  \city{St. Gallen}
  \country{Switzerland}
\email{kayerik.jenss@gmail.com}
}
\author{Christian Hildebrand}
\affiliation{%
  \institution{Institute of Behavioral Science and Technology, University of St. Gallen}
  \city{St. Gallen}
  \country{Switzerland}
\email{christian.hildebrand@unisg.ch}
}

\renewcommand{\shortauthors}{Hostettler et al.}

\begin{abstract}
Industrial robots become increasingly prevalent, resulting in a growing need for intuitive, comforting human-robot collaboration. We present a user-aware robotic system that adapts to operator behavior in real time while non-intrusively monitoring physiological signals to create a more responsive and empathetic environment. Our prototype dynamically adjusts robot speed and movement patterns while measuring operator pupil dilation and proximity. Our user study compares this adaptive system to a non-adaptive counterpart, and demonstrates that the adaptive system significantly reduces both perceived and physiologically measured cognitive load while enhancing usability. Participants reported increased feelings of comfort, safety, trust, and a stronger sense of collaboration when working with the adaptive robot. This highlights the potential of integrating real-time physiological data into human-robot interaction paradigms. This novel approach creates more intuitive and collaborative industrial environments where robots effectively 'read' and respond to human cognitive states, and we feature all data and code for future use.
\end{abstract}



\maketitle

\section{Introduction}\label{sec1}

The increasing presence of robots in industrial settings is fundamentally changing the nature of human work. As these machines become more sophisticated and ubiquitous, there is a growing need for intuitive and adaptive human-robot interaction (HRI) systems that can respond to the cognitive and emotional states of human operators in real time. This paper presents a novel approach to industrial HRI that utilizes physiological signals, particularly pupil dilation and spatial distance between operator and robot, to create a more responsive, adaptive robotic system and to monitor the success of such adaptation. The current work builds on a stream of research on\textit{ intelligent user interfaces} that improve the interactions between users and machines ~\cite{Arazy, Alvarez} based on an integrative approach to user modeling, user adaptivity, and personalization toward the user \cite{Alvarez}. The rapidly increasing accessible amount of data enables and requires \emph{intelligent adaptive systems} that sense operators' and environmental states and respond appropriately, resulting in more efficient, effective and safer interactions that ``allow operators a more productive and rewarding life''~\cite{Hou}. As the increasing number of robots deployed in industry~\cite{IFR} affects thousands of operators in their daily work routine, we transfer these key principles of intelligent adaptive systems to human-robot interaction (HRI), and in particular to articulated robot arms. 

On manufacturing shopfloors, robots support productivity and employees are relieved of monotonous and physically demanding tasks. However, new issues arise with regards to the acceptance of these new ``colleagues'' and the design of ergonomic, safe, and enjoyable industrial HRI. With the rising number of collaborative robots (cobots) that require fewer safety precautions, interactions and collaboration between humans and robots become spatially closer, reinforcing the need to ensure physiological as well as psychological safety and well-being. This is reflected already in industry, where a study among 200 Italian companies reports that 68\% have a demand for monitoring workers' interactions with machines \cite{PERUZZINI2017806}. We see such human-robot ensembles as socio-technical cyber-physical systems (STCPS), and hence ``less as a dichotomy but rather as human and nonhuman actors in a sociotechnical network that needs to be designed as such''~\cite{Weiss2021}. In addition, several researchers describe the evolution from co-existence of robots and operators towards human-robot collaboration (HRC) \cite{SIMOES202228, Matheson}, HRC causing higher workcell complexity with regard to topics such as dynamic task allocation and sophisticated sensing abilities of the human-robot team. STCPSs that sense the state of their interaction partners and context, and adapt their movement behavior in real time might enable improved industrial HRC to ensure more ergonomic, enjoyable interactions, create a better user experience, and ultimately increase robot acceptance. 
To realize such adaptive STCPS for industrial robots, we lack a deeper understanding of (1) what information about its situation an STCPS needs to sense for appropriate adaptation, (2) the specific, practical movement adaptations for articulated robots, and (3) how the sensed information should be connected or mapped to these adaptations.

To detect relevant and appropriate human characteristics, to design practicable robot movements, and to connect them effectively, requires a foundational understanding of how these components interrelate, and how they can be combined to achieve improved interactions with robots. Some of these aspects have received attention in HRI research, and we offer a brief overview on existing findings in Section~\ref{sec2}. However, the causal relationships between robotic actions and human responses remain fragmented, and human evaluations of a comprehensive adaptive robot system have not been examined either conceptually nor as part of an empirical demonstration yet. With our work, we contribute to the existing literature in three important ways: 

\begin{enumerate}

 \item First, we present the first industrial robot system that adapts its behavior in real time based on operator spatial proximity while collecting pupil dilation data as evaluative feedback. This system represents a significant advancement in creating more intuitive and responsive HRC. We provide a detailed description of the system's architecture, its ability to track both subjective and objective human responses, and make our source code openly available for further development and replication studies. 
 
 \item Second, we conduct a comprehensive user study that demonstrates the effectiveness of our adaptive system. Our results show significant improvements in both subjective measures (lower reported cognitive workload, improved usability, greater perceived subjective safety, and higher trust) and objective physiological indicators (pupil dilation patterns indicative of reduced cognitive load and lower stress levels). This study provides empirical evidence for the benefits of integrating real-time adaptive behavior in an industrial robot system, and might generalize far beyond.
 
 \item Finally, as a broader methodological contribution, we introduce a new methodological approach for analyzing complex, high-dimensional datasets generated from adaptive human-robot interactions. This framework can serve as a blueprint for evaluating and benchmarking future STCPS, enabling more standardized comparisons across different systems and contexts.

\end{enumerate}

Section~\ref{sec2} links the general idea of situation or context awareness and system adaptiveness to the current state of research in HRI. We derive and discuss system components that the system uses to adapt its behavior and that form the foundation for our adaptive robot system. Using a practicable selection of subjective and objective human factors and robot behaviors that can be deployed with an articulated industrial robot, we introduce our implemented prototype of an adaptive robot STCPS. In Section~\ref{sec3}, we present our experimental setup and methodology to assess the effectiveness of the robotic system, and report the empirical results of our user study. We broaden our perspective in Section~\ref{Sec4} and discuss our findings with regard to implications and recommendations for future adaptive user systems and related HRI research.

\section{Related Work and Theoretical Foundations}\label{sec2}

When interacting with industrial robots, the necessity to ensure human safety has led to many technological advancements including safe robot control and motion planning as well as prediction and recognition of human and robot actions. These approaches ensure that robots do not physically harm humans and refer to \emph{physical safety}. On the other hand, robots may violate social conventions and norms during interaction that affect operators' psychological states and trigger discomfort or stress, which corresponds to \emph{psychological safety}~\cite{Lasota2017, Story2022} and partly to \emph{mental health or well-being} as described in~\cite{EU-project}. To ensure physical safety, most industrial robots already incorporate either protective force stops or other safeguards that increase the robot's awareness of its surroundings~\cite{Kumar2017}. However, psychological safety necessitates additional sensing abilities in regards to context- and human-awareness. More aware robot systems might support stress reduction and increase ergonomics, and possibly can reinforce further favorable outcomes such as human acceptance, trust, likability, and enjoyment of the respective robot, in line with demands that have been articulated for human-automation symbiosis in~\cite{Romero}. 

\subsection{Ergonomic Robot Behavior}

The selection of adaptable robot-related factors of an articulated robot arm that influence human perception is limited. Findings that consider appearance and behavior of humanoid robots (which are plenty) are only of limited use when considering \emph{articulated} robotic arms. The operational context of industrial robots discourages the addition of humanoid features (e.g., a face or eyes) to industrial robots, and behavioral adaptations in this context can be implemented by modulating movements only. In addition, organizations that use industrial robots strive for productivity gains, which further affects the suitability of possible motion variations. 
Several studies in HRI research focus on transparency in terms of predictability and legibility of robot motions~\cite{Kuz2012ThePrediction, Brecher2013TowardsRobots, Kuz2013UsingI, Dragan2013LegibilityMotion} as well as ergonomic hand-over motions~\cite{Mainprice2010PlanningInteraction, Bestick2018LearningHandovers, Dehais2011PhysiologicalTask}, and previous findings in this field have demonstrated that human perceptions and evaluations of industrial robots depend on movement behaviors~\cite{Hostettler2022}. The adaptation of movement parameters (such as speed or smoothness) to human preferences hence may lead to favorable evaluations of HRI, such as stress reduction, increased ergonomics and enjoyment. 
\cite{Lasota2017} and~\cite{Story2022} provide extensive overviews of the effects of robot speed and proxemics behavior on various physiological safety metrics and found relations between speed and workload~\cite{Story2022}; they further highlight the complexity resulting from interactions between robotic actions and human evaluations, such as the dependence of optimal robot speed on outside aspects that cannot currently be adapted at run time (e.g., the robot's size or the human's level of prior experience with robots; cf.~\cite{Lasota2017}). \cite{Story2022} found significant positive relationship between a UR5 robot's speed and participants' cognitive workload (CWL), even though the effect was only significant above a certain threshold. However, they found additional factors; for instance, with increasing robot speed, participants felt that they needed to complete the task faster to keep up with the robot, which may be an additional source of increased CWL. This highlights another difficulty, namely to detect actual relationships between sensed human and context factors, movement behaviors, and affiliated confounding variables when testing the system. Regarding proximity, \cite{Story2022} found no effect on CWL even though previous studies did report significant effects. Again, certain proximity thresholds as well as the participants being in control of the proximity (e.g., by remaining free to increase their distance to the robot) may explain the differing results, and have to be taken into account when testing the system. \cite{Eimontaite2020} found that robot speed and synchronization between the human and the robot is required to achieve workforce satisfaction and well-being. Corresponding to~\cite{Hostettler2023}, an industrial robot's speed has been shown to influence CWL as indicated by pupil dilation and can be modulated to increase preference. This points to a key feature of real-time adaptiveness and its suitability for practical industrial contexts: the usage of \emph{non-intrusive} measurement methods of relevant factors that trigger certain robot behaviors. 

\paragraph{Applicable Movement Modulations for Articulated Robot Arms}

Summarizing, existing findings suggest that movement modulations based on robot speed, smoothness of movements, and movement range represent robot-related modulations that can be adapted to human preferences, and that adaptation of these movements might have a positive effect on the human user. Building on previous studies \cite{Story2022, Eimontaite2020, Lasota2017, Hostettler2022}, we focus on robot movement range implementable as distance to the user and speed as movement parameters in our system prototype. This goes in line with general findings on industrial manipulators presented in~\cite{Rubagotti}, according to which larger human-robot distances as well as low robot speeds increase the feeling of perceived safety. Regarding smoothness of movements that has been shown to affect perceived human-likeness according to~\cite{Hostettler2022}, initial tests have demonstrated that programming the rounding of curves with UR robots as described in~\cite{Hostettler2022} is not directly transferable to the robot available in our lab, which is a UFactory's xArm7, and we therefore ignore smoothness as additional movement parameter.

\subsection{Adaptive Behavior Triggers}\label{HumanFactors}

\paragraph{Static Human-Related Factors}

Regarding human-related factors, only few insights about which individual differences actually determine the subjective human perception of HRI exist, such as findings on gender differences in~\cite{Bishop2019SocialAcceptance,Abel2020GenderActions}, on attitude and familiarity with robots in~\cite{Szczepanowski2020EducationRobots}, and on the effect of prior experience with robots on trust in~\cite{Sanders}. However, most of these findings relate to humanoid or social robots. Some of the findings are also inconsistent: for example, \cite{Kuo} found only limited relevance of age when interacting with a Peoplebot, and males having more positive attitudes towards the robot than females. Conversely, \cite{wagner} reports that females evaluate their interaction with robots as more useful and satisfying, but confirmation for a more positive attitude towards the robot by males, and \cite{Graaf} finds older people to be more likely to enjoy using service robots despite a lower intention to use them. Given the static characteristics of most of these factors and our objective to implement real-time adaptiveness, we require alternative factors and technologies that allow to capture dynamic user responses during interactions.

\paragraph{Dynamic Human-Related Factors}

Recent research has demonstrated that physiological and behavioral measures are associated with distinct stress responses and safety perceptions during interactions with robots, such as heart rate (HR) and skin potential response (SPR)~\cite{Kühnlenz, Lasota2017}, pupil dilation~\cite{Hostettler2023} and proxemics behavior~\cite{Walters}. Similarly, to monitor user experience (UX) with Industry 4.0 applications,~\cite{PERUZZINI2017806} proposes a framework that includes several physical and physiological measurements such as heart rate variability (HRV), electro-dermal activity (EDA), galvanic skin response (GSR), and body activity, but the actual relations between these measurements and resulting system instructions remain unclear.

\paragraph{Suitability for Real-Time Data Collection}

Real-time reactiveness requires feedback loops that sense static as well as dynamic human states and behaviors, and respond with appropriate movement adaptions. To detect and ensure the appropriateness of robot behavior adaptions, three tools can be used: questionnaires, physiological metrics, and behavioral metrics. These human responses differ with regard to their suitability for real-time detection of human states, the objectiveness of the response, and their intrusiveness for the human~\cite{Lasota2017}. To implement the overall idea of real-time awareness and reactiveness, we propose to focus on objective metrics that are detectable at run time, rather than static individual differences or questionnaires. Still, to identify the relevant metrics, we need an understanding of how objective metrics in our setting correlate to existing, self-reported items such as the connection between self-reported CWL and pupil dilation. For this reason, our user study includes questionnaires with the objective to correlate questionnaire responses with objective, dynamically measurable data; we then propose to replace self-reports if appropriate correlations can be shown.

\paragraph{Suitability for Non-Intrusive Data Collection}

Moreover, to build a system that might be utilizable in actual shopfloors, we focus on metrics that can be captured preferably in a non-intrusive manner. Even though there has been increasing interest and research on using physiological measures to capture human states, the topic remains underinvestigated in industrial HRI~\cite{Gervasi2022, Bethel}. \cite{Rani2004} presents an affect-sensitive system that detects human anxiety using physiological measurements such as cardiac, electrodermal, and electromyographic responses, and \cite{kulic} compares self-reported responses to different robot speeds and trajectories with skin conductance and heart rate measurements. \cite{Eimontaite2020} shows that the presence of an operational robot in a collaborative task leads to increased arousal levels based on skin conductance. \cite{Story2022} shows that workload correlates with the speed of an industrial robot arm, even though \cite{CHARLES2019221} provide a critical analysis of physiological measures' ability to measure mental workload while, using a mobile eye-tracker, pupil dilation as an indication of CWL has been shown to be sensitive to robot speed \cite{Hostettler2023}. Indeed, the relation of cognitive processing and pupil dilation has been investigated for decades~\cite{Beatty1982TaskevokedPR}, and pupil diameter has been shown to differentiate workload between task types and to be an appropriate measure applicable to task durations of < 5mins~\cite{CHARLES2019221}. Besides these stress-related human responses to robotic actions, safe interactions with robots are crucial in the industrial context, where operators collaborate with robots on a daily basis. Accordingly, \cite{Rani2004} describes the idea of detecting human fatigue in manufacturing shopfloors, and robots taking appropriate actions as safety precautions. \cite{Urasaki} used blink frequency to measure human fatigue as a result of differing robot motion timing. With regard to physical safety,~\cite{Mohammed} presents how to avoid collisions in an augmented environment using a depth camera,~\cite{MAGRINI2020101846} combines Microsoft Kinect sensors and laser scanners to ensure operator safety, and~\cite{long} uses exteroceptive sensors to adapt robot speed based on operators' presence.

\paragraph{Context-related Factors}

Finally, cultural contexts have been shown to result in differing HRI experiences~\cite{Salem2014, Trovato2013} and therefore might be taken into account as well. The robot's context observed through its location as an approximation to cultural norms it has to comply with might allow to implement variations in the robot's movement adaptions that fit cultural preferences to its environment.

\paragraph{Resulting Factors Suitable for Real-Time Adaptiveness in Manufacturing Shopfloors}

To sufficiently take the organizational real-life environment into account and based on the demonstrated relevance of the above-mentioned factors in HRI, we focus on physiological and behavioral metrics that are as objective as possible, detectable as unintrusively as possible, and in real time. Combined with existing findings on the influence of individual differences presented above, Table~\ref{tab:measurements} lists a selection of such factors that we have considered, tested and evaluated, complemented by a possible measurement device and its suitability for real-time applications. In addition, we list existing exemplary work that has used similar concepts in the HRI field.

\begin{table*}[t]
\caption{Human- and Context Factors, Measurement Concepts, Measurement Devices, and Exemplary Work}
    \label{tab:measurements}
\Description{The table contains a list of human- and context factors that can potentially be used to trigger system adaptations. The rows list 9 subjects: stress, CWL, fatigue, proxemics behavior, emotional reaction, prior experience, user experience, cultural norms and other environmental conditions. The 5 columns list their properties: measurement concepts, measurement device, temporal properties, ascertainability in real-time and exemplary work.}
\centering
\resizebox{\textwidth}{!}{%
\begin{tabular}{llllll}
\textbf{Subject}                                                          & \textbf{Measurement Concepts}                                                                     & \textbf{Measurement Device}                                                             & \textbf{Temporal Properties}                              & \textbf{\begin{tabular}[c]{@{}l@{}}Ascertainability in \\ Real-Time\end{tabular}} & \textbf{Examplary Work}                                                                                                                                                                             \\ \hline
Stress                                                                    & \begin{tabular}[c]{@{}l@{}}Heart Rate Variability \\ (HRV)\end{tabular}                           & Apple Watch                                                                             & Dynamic                                                   & Yes                                                                               & \begin{tabular}[c]{@{}l@{}}Velocity profiles and HRV \\ \cite{Kühnlenz}\end{tabular}                                                                                               \\ \hline
\begin{tabular}[c]{@{}l@{}}Cognitive Workload \\ (CWL)\end{tabular}       & \begin{tabular}[c]{@{}l@{}}Pupil Dilation\\ Questionnaire\end{tabular}                            & \begin{tabular}[c]{@{}l@{}}Pupil Core Eyetracker\\ NASA-TLX questionnaire\end{tabular}  & \begin{tabular}[c]{@{}l@{}}Dynamic\\ Dynamic\end{tabular} & \begin{tabular}[c]{@{}l@{}}Yes\\ No\end{tabular}                                  & \begin{tabular}[c]{@{}l@{}}Robot speed and pupil dilation \\ \cite{Hostettler2023}\\ Collaborative assembly tasks \\ and NASA-TLX \cite{Su}\end{tabular}          \\ \hline
Fatigue                                                                   & Blink Rate                                                                                        & Pupil Core Eyetracker                                                                   & Dynamic                                                   & Yes                                                                               & \begin{tabular}[c]{@{}l@{}}Motion timing and blink \\ frequency \cite{Urasaki}\end{tabular}                                                                                        \\ \hline
Proxemics Behavior                                                        & \begin{tabular}[c]{@{}l@{}}Distance user / Tool-Center \\ Point (TCP)\end{tabular}                & Intel Real Sense Camera                                                                 & Dynamic                                                   & Yes                                                                               & \begin{tabular}[c]{@{}l@{}}Likeability, gaze behavior and \\ proxemics \cite{Mumm}\end{tabular}                                                                                    \\ \hline
Emotional Reaction                                                        & \begin{tabular}[c]{@{}l@{}}Facial Expression Analysis \\ (FEA)\end{tabular}                       & \begin{tabular}[c]{@{}l@{}}Intel Real Sense Camera \& \\ Deepface Software\end{tabular} & Dynamic                                                   & Yes                                                                               & \begin{tabular}[c]{@{}l@{}}Robot ignoring, mirroring or \\ displaying human facial \\ expressions \cite{Gonsior}\end{tabular}                                                      \\ \hline
Prior Experience                                                          & \begin{tabular}[c]{@{}l@{}}Interests stated on platforms \\ like Linkedin or Twitter\end{tabular} & Data Scraping                                                                           & Static                                                    & Yes                                                                               & \begin{tabular}[c]{@{}l@{}}Prior interaction with robots and \\ attitudes and trust \cite{Sanders}\end{tabular}                                                                    \\ \hline
User Experience                                                           & Questionnaire                                                                                     & User Experience Questionnaire                                                           & Dynamic                                                   & No                                                                                & UX design tools in HRI \cite{Prati}                                                                                                                                                \\ \hline
Cultural Norms                                                            & Location of the system                                                                            & IP                                                                                      & Static                                                    & Yes                                                                               & \begin{tabular}[c]{@{}l@{}}National culture and attitude \\ towards robots \cite{Evers}\end{tabular}                                                                               \\ \hline
\begin{tabular}[c]{@{}l@{}}Other Environmental \\ Conditions\end{tabular} & \begin{tabular}[c]{@{}l@{}}Time of day, shift duration, \\ lightning conditions\end{tabular}      & Various                                                                                 & Dynamic \& Static                                         & Yes                                                                               & \begin{tabular}[c]{@{}l@{}}Human detection and action \\ recognition for human safety \cite{Bonci}\\ Pupil size and lighting sources \cite{Rossi}\end{tabular} \\ \hline
\end{tabular}%
}
\end{table*}

Based on pilot testing, we select two dynamic physiological and behavioral factors that can be collected in real time and in a minimally invasive way for our prototype: \emph{pupil dilation} and \emph{proxemics} behavior. We give a brief account of why we did not select the other factors in the following: We tested \emph{HRV} with hyperventilating as an approximation to stress situations; however, our tests revealed that the cool-down period after increased stress levels was 1-2 minutes which makes HRV unsuitable for real-time reactiveness. The test setup was based on an Apple Watch, which was worn on the participants wrist, which is the least intrusive way to collect this data in a potential shopfloor scenario, but provides significantly lower data accuracy than a chest-mounted sensor or an electrocardiogram. 
We further considered \emph{fatigue} detection based on \emph{blink rate}; however, our experiments are conducted in rather short time frames and fatigue did not occur within these time frames (even though it was replicable by blinking intentionally). We further tested the effect of a humorous robotic behavior---the robot unexpectedly grips and honks a horn---on \emph{expressed emotions}, and found a strong effect on the expression of joy. However, to remain focused on the organizational context where industrial robots are used, we do not consider this behavior in our system prototype. Regarding \emph{prior experience} with robots, we have tested data scraping of user Linkedin profiles, focusing on indications regarding experience with an interest for industrial equipment, manufacturing processes, or robots in general. Based on the participants listed skills, education, professional experience, as well as certifications and publications, the data was fed into a large language model (OpenAI GPT-3.5 Turbo), attempting to transfer these qualitative measures of the operator into a quantitative score on a predefined scale. Even after tuning the input prompts, our model was not able to output meaningful values, as experienced as well as inexperienced individuals received very similar scores. Regarding context factors, we expect cultural dependencies to be relevant as indicated in~\cite{Salem2014, Trovato2013}. Since our user study is not conducted in diverse cultural environments, we do not include culture-dependent movement behaviors. However, cultural context may be included by adapting adaptation characteristics themselves (e.g., the strength of behavioral changes or setting different initial speed levels according) to cultural preferences. In addition, environmental conditions such as time of day as an indication of an operator's alertness based on how long a shift already lasts, or lighting condition to classify the informative power of pupil dilation are conceivable. However, these factors require knowledge of additional causal relationships that are so far partly unexplored, and to keep our prototype simple and testable we ignore these additional factors. Consequently, we focus on proxemics behavior and pupil dilation that have both received much attention in previous research, and fulfill the needs required for real-time detection, objectiveness, and non-intrusiveness. To acquire additional knowledge on how these objective factors relate to self-reported human preferences, we collect perceived CWL and further data on the users' experience using questionnaires.

\section{Evaluation of an Adaptive Industrial Robot} \label{sec3}

\begin{figure*}[t]
    \centering
    \includegraphics[width=10cm]{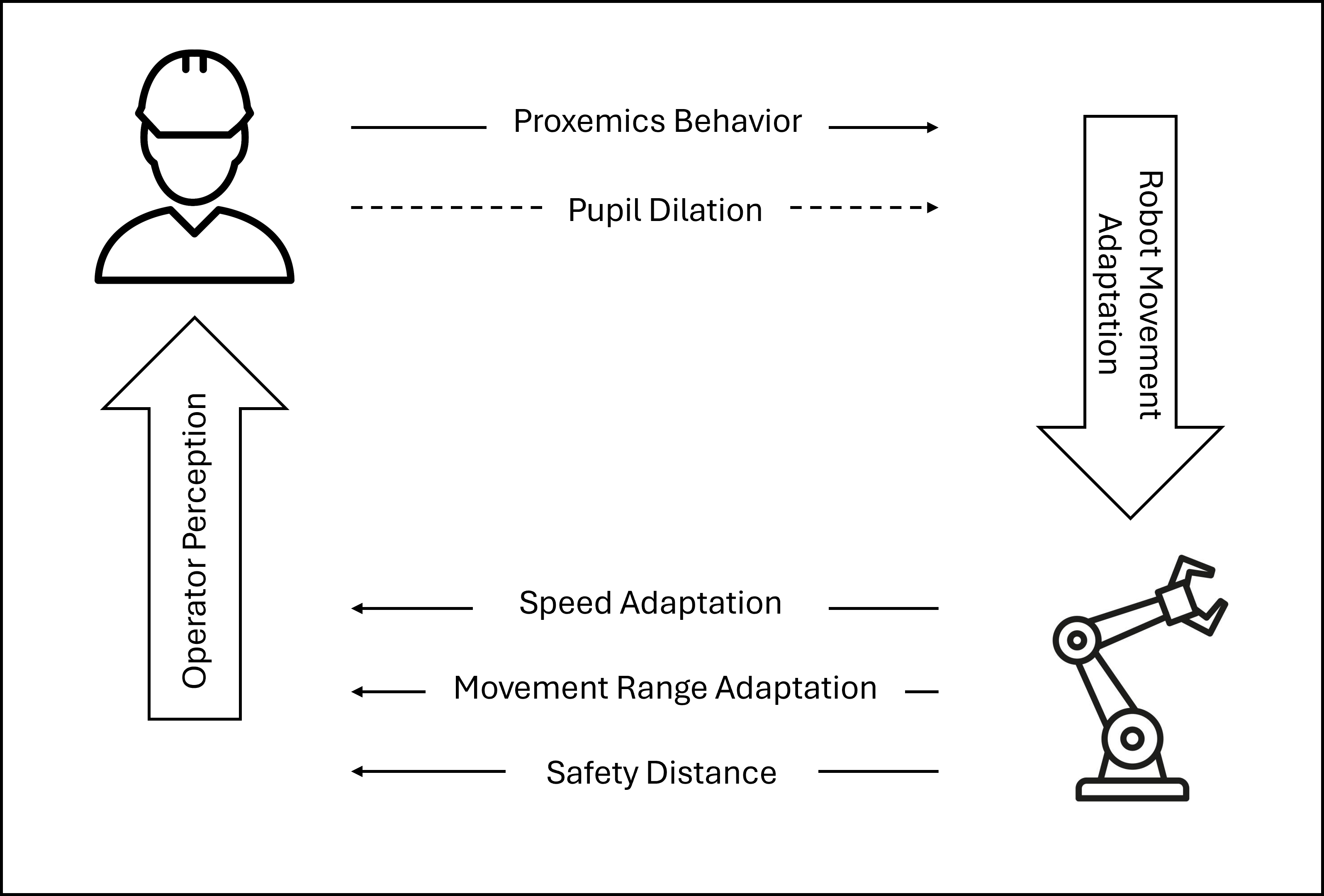}
    \caption{System Architecture Illustrating the Perceived Operator's States and the Robot's Movement Adaptations.}
    \Description{The figure depicts a feedback loop where the operator perception triggers robot movement adaptations. Operator perception includes proxemics behavior and pupil dilation, and robot movement adaptations include speed and movement range adaptation as well as keeping a safety distance.}
    \label{fig:2}
\end{figure*}

Following existing findings and the discussed restrictions in Section~\ref{sec2}, we implemented a real-time adaptive industrial robot system based on a uFactory xArm7 articulated robotic arm as illustrated in Figure~\ref{fig:2}\footnote{The system's source code is available in our data repository on OSF, see \url{https://osf.io/4kgsa/?view_only=76da10728dc94d339ccbebb9ebbda48e}}. In our study setup, we record the user's proxemics behavior using an Intel RealSense D455 depth camera; our system computes the distance between the user's torso and the robot's center (with a low-pass filter to account for jitter in the detection of the human torso), and additional distance measurements (e.g., user torso to robot tool center point) are done using an additional ceiling-mounted camera. To include stress- and workload-related responses, we measure pupil dilation using a Pupil Labs Core eye tracker. We record pupil dilation data to analyze its correspondence with self-reported CWL data, as well as how the robot's adaptiveness affects the users' workload and how distance between the robot and the user affects pupil dilation, building the foundation for future systems that adapt robot movements to CWL measured through real-time pupil dilation. Since we do not know yet how interacting with our robot system affects proxemics behavior and pupil dilation, and to ensure that we do not mix up the effect of these factors, we adapt robot behavior to the distance between the user and the robot only. We furthermore use the raw NASA Task Load Index (NASA-TLX) questionnaire\footnote{See \url{https://humansystems.arc.nasa.gov/groups/tlx/downloads/TLXScale.pdf}} to collect self-reported data on CWL. Focusing on ``how a person feels about using a product, i.e., the experiential, affective, meaningful and valuable aspects of product use''~\cite{Vermeeren}, we use the short version User-Experience-Questionnaire (UEQ-S)\footnote{See \url{https://www.ueq-online.org/}}. The real-time adaptation of our robot follows three assumptions:

\begin{enumerate}

\item Lower movement speed of the robot in general leads to higher human preference, and vice versa. Speed reduction reduces stress/CWL.

\item Interacting with the robot in close proximity is less preferred (and increases stress/CWL), and preference increases (and decreases stress/CWL) if distance is increased. High movement speeds in close proximity are less preferred (and increase stress/CWL), and speed reduction in close proximity is preferred (and decreases stress/CWL).

\item Interactions in very close proximity are less preferred (and increase stress/CWL), and ensuring a safety distance between the user and the robot's TCP increases preference (and decreases stress/CWL).

\end{enumerate}

In our system, these assumptions are implemented as per the adaptation specifications in Table~\ref{AdaptionSpecifications}. 

\begin{table*}[]\tiny
 \caption{Adaptation Specifications}
\label{AdaptionSpecifications}
\Description{The table explains the used adaptation specifications, where speed, proximity and speed, and safety distance are the rows, and adaptation mode and specifications are the columns. Speed is adapted in four speed levels (30\%, 50\%, 80\% and 100\%. Proximity and speed is considered with speed reductions at four proximity levels, namely 30cm, 60cm, 100cm and above 100cm from the robot's center. Safety Distance is applied if the user stands closer than 40cm to the robot, and avoids collisions through runaround.}
\centering
\resizebox{\textwidth}{!}{%
\begin{tabular}{lll}
\textbf{Parameter} & \textbf{Adaption Mode}                                                                                                                                              & \textbf{Specifications}                                                                                                         \\ \hline
Speed              & Adaptation of movement speed                                                                                                                                          & \begin{tabular}[c]{@{}l@{}}4 speed levels: 30\%, 50\%, 80\% and 100\%  of \\ the robot's task-specific maximum speed level\end{tabular}\\ \hline
Proximity + Speed  & \begin{tabular}[c]{@{}l@{}}Distances between the robot and the user \\ trigger speed reduction/increase\end{tabular}                                                & \begin{tabular}[c]{@{}l@{}}4 proximity levels: 30cm, 60cm, 100cm and\\ \textgreater{}100cm from the robot's center\end{tabular} \\ \hline
Safety Distance    & \begin{tabular}[c]{@{}l@{}}Very close proximity to the robot leads to \\ changed trajectory to ensure that the robot \\ does not collide with the user\end{tabular} & \begin{tabular}[c]{@{}l@{}}Gradual avoidance of collisions and \\ runaround if proximity is below 40cm\end{tabular}\\ \hline
\end{tabular}%
}
\end{table*}

\subsection{Study Design}\label{Sec3.1}

To test the effectiveness of the implemented adaptiveness functionalities, we implemented two versions of the system: An adaptive version according to the above-mentioned adaptation specification and a non-adaptive version where the robot's movement range, trajectory, and speed remain unchanged, where the speed is given by the task-applicable maximum. Even though we believe that our system contributes to the development of HRC as described above, our test rather focuses on a co-existence task in a first step, but includes a handover situation to include collaborative elements. Observing the robot rather than directly collaborating with it goes in line with most existing robots in shopfloors that work behind fences, and therefore complies with actual robot existence that affects today's robot users in industry. In addition, the reduction of complexity allows to understand basic human responses to robots that work in close proximity to operators which is a common case for cobots without safety fences. Regarding the vulnerability of pupil diameter measurements related to lightning conditions, we have ensured similar lighting conditions for all participants by closing all blinds of the lab floor and turning on the ceiling lights on the highest level. We carried out a within-subject user study with 16 participants (12 male, 4 female). Participants were recruited at the authors' institution and received the institution's standard financial compensation for participation. The experiment took around 20 minutes per participant. The mean age of participants was 31.63 years (SD = 8.72), ranging from 24 to 58 years old. Regarding educational background, 7 participants (43.75\%) held a Bachelor's degree, 5 (31.25\%) held a Master's degree, and 4 (25\%) held a PhD. Participants reported varying levels of prior experience with robots, ranging from 1 (very low) to 10 (very high) on a 10-point scale. The median level of prior experience was 3, with four participants each reporting levels 2 and 3, three participants reporting a level of 5, and one participant reporting a level of 10.
Technical affinity was measured on a 10-point scale, with values ranging from 6 to 10. The median technical affinity level was 8, with 5 participants reporting this level. Four participants reported a technical affinity of 9, while three participants reported a value of 7 and two reported a level of 10.

To investigate the effect of the system's adaptiveness on human responses, we used subjective (self-reported NASA-TLX questionnaire and UEQ-S) and objective (recorded pupil dilation data and proximity behavior) measures. We hence analyzed the following relationships:

\begin{enumerate}

    \item Pupil diameter and distance to the robot.

    \item Pupil diameter and robot adaptiveness.

    \item Self-reported CWL and pupil diameter.

    \item Self-reported CWL and robot adaptiveness.

    \item Pupil diameter and critical, close-proximity situations.
  
    \item Usability and robot adaptiveness.

    \item Qualitative perception and robot adaptiveness.
             
\end{enumerate}

\subsection{Experimental Procedure} \label{ExperimentalProcedure}

\begin{figure*}[t]
    \centering
    \includegraphics[width=15cm]{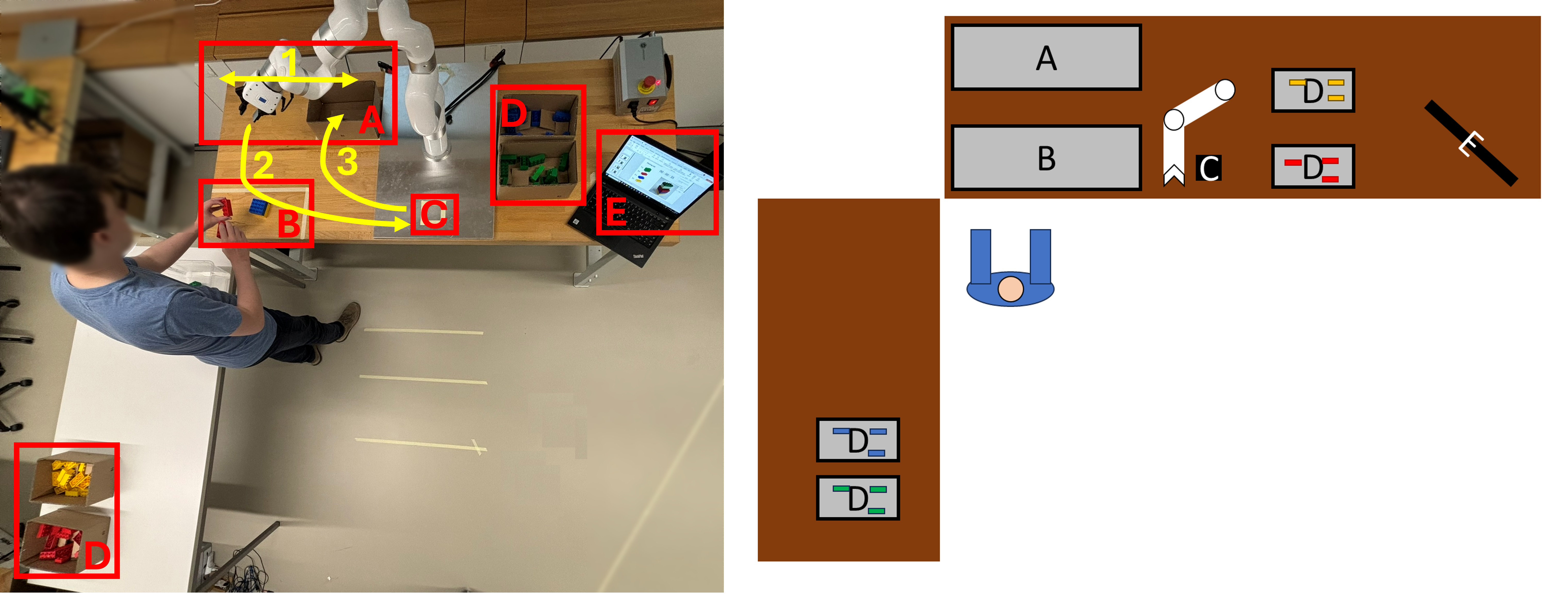}
    \caption{Experimental Setup Showing the Movement Episodes 1-3 and the Areas A-E Further Described in the Text.}
    \label{Setup}
    \Description{The figure depicts the experimental setup with the robot mounted on a table, the user standing in front of the robot and the containers with Lego parts placed to the right of the robot, as well as on an additional table on the left.}
\end{figure*}

Participants were briefly instructed about the experimental procedure. We explained the robot's adaptiveness functionalities, answered open questions and asked all participants to sign an informed consent form in the beginning of the experiment. The study procedure consisted of two parts: a \emph{Trial Task} and an \emph{Assembly Task}. In the \emph{Trial Task}, participants were introduced to the robot systems' functionalities and interacted with the robot freely in both, the adaptive (Trial 1) and the non-adaptive (Trial 2) condition. In the \emph{Assembly Task} of the study, participants were given the task to assemble a Lego model according to a visual instruction and hand it over to the robot, again, in the adaptive (Assembly 1) and non-adaptive (Assembly 2) condition. The order of the conditions was randomized in both, the Trial Task and the Assembly Task. Figure~\ref{Setup} illustrates the experimental setup. In both parts (Trial/Assembly),  the robot performed three episodes. In episode 1, the robot ``worked'' independently in area (A). We refer to this episode as \emph{works independently} in the following. In episode 2, the robot moved from area (A) through area (B) to the pick-up place (C), and we refer to this episode as \emph{forward}. In episode 3, in the following referred to as \emph{backward}, the robot moved back to area (A). 

In the first part of the experiment (Trial 1/2), the robot performed this procedure (Episodes 1-3) repeatedly. Participants were free to test the robot's functionalities by getting closer and increasing their distance to the robot in both, the adaptive and in the non-adaptive condition, for as long as they wanted. In this way, we ensured that participants were aware of the robot's adaptive mode, while a pilot study had previously shown that starting with the main experiment immediately overloaded participants and specifically resulted in them not perceiving the robot's adaptiveness in the first few interactions. Giving participants the opportunity to get to know the system's functionalities might reduce the surprise effect and some of the unconscious, implicit reactions; however, this corresponds much better with an actual robot/operator situation where the operator works with a robot on a regular basis. The trial phase took 1 to 5 minutes per participants, until the participants signaled that they were ready for the next part of the experiment. To remain able to analyze human reactions even during the trials, we already collected proximity and pupil dilation data during this introduction and this also serves as manipulation check (see the results discussion in Section~\ref{Sec4}). 

In the second part (Assembly 1/2), the robot only performed this procedure (Episodes 1-3) once per condition. Participants were shown a picture of a Lego model on a screen (E, see Figure~\ref{Setup}), then needed to pick the required Lego bricks from four different containers (D), assemble them correctly in a specified area close to the robot (B), and hand them over to the robot (C). While a participant worked on these tasks, the robot was moving in area (A) and moved to (C) every 35 seconds to take the assembled Lego model from the operator. This resembles an assembly task in a production line next to a robot, where the 35s-timebox was defined based on pre-tests, ensuring all participants have enough time to finish the task. As the robot reduces speed in close proximity to the user, and accordingly may take more time to move from (A) to (C), we have incorporated a time buffer at (C) to ensure the robot picks the finished part exactly after 35s in both conditions. Regarding the Lego model complexity, all models were pre-tested to ensure similar complexity levels. We used simple models and only varied the arrangement of colors per model, to account for workload that might otherwise differ between the Lego models, or higher workload as a result of the assembly complexity of the Lego model itself. The arrangement of the assembly area (B) and the four containers (D) ensure that participants worked in close proximity to the robot, but always had the option to step back and increase the distance to the robot if they felt uncomfortable---this is, again, modeled according to actual assembly tasks in a production line. A timer on screen (E) as well as an acoustic signal indicated the remaining time. The Assembly Task took 45 seconds per condition. Again, we collected proximity and pupil dilation data during the entire interaction. 

In addition to the objective human reactions to the interaction with the robot (distance and pupil data), we collected several self-reported measurements using questionnaires. To measure CWL during the assembly task, we asked the participants to fill out the raw NASA-TLX questionnaire and the UEQ-S after Assembly 1 and Assembly 2, as a self-reported indication of the users' perceived workload and experience. To additionally compare the two conditions in the users' words, we asked several open questions regarding perceived differences between the conditions and their general perception at the end of the experiment.

The entire experiment took between 10 and 20 minutes per participant, depending on the length of the trial task and the time it took participants to fill out the questionnaires and answer the open questions.

\subsection{Data Processing and Analysis}

While robot-related distance and speed data were logged by our system, pupil diameter data were exported using the Pupil Labs software\footnote{See \url{https://pupil-labs.com/products/core}}, preprocessed and aggregated for analysis. We matched all data streams based on the exported (synchronized) timestamps and removed actual zero values (e.g., non-tracked eyes) and averaged pupil diameter measurements across both eyes per participant. The data were then sorted by participant, episode (works independently, forward, backward), and step (Trial 1/2 \& Assembly 1/2). A common~\cite{Campbell, VanEgroo} standard deviation filter was applied to remove outliers, excluding data points outside three standard deviations above and below the mean diameter for each participant/episode/step combination.
The cleaned data were then aggregated by participant, episode, step, and adaptiveness condition. For each of these groups, we calculated several summary statistics: mean distance, mean pupil diameter, first quartile (Q1) of pupil diameter, third quartile (Q3) of pupil diameter, and the number of observations (N).

To ensure comparability among variables that were initially on different scales, we standardized the distance variables as well as the number of observations (N). Standardization was performed using the \texttt{scale()} function in R, transforming each variable into a mean of zero and a standard deviation of one. This approach facilitates uniform scaling across variables with different measurement units, enabling a more straightforward interpretation and comparison of the results. Finally, the NASA-TLX questionnaire data---measured separately for steps `Assembly 1' and `Assembly 2' in the adaptive and non-adaptive conditions---were incorporated into the analysis. All datasets and all analysis scripts are available in our data repository on OSF\footnote{See \url{https://osf.io/4kgsa/?view_only=76da10728dc94d339ccbebb9ebbda48e}}.

To assess relationships (1), (2), and (3) from Section~\ref{Sec3.1}, we first estimated a linear mixed-effects model using the \texttt{lme4} package~\cite{Bates} in R to examine the \emph{relationships between pupil diameter and our set of key predictors}. The dependent variable was mean pupil diameter. Fixed effects included mean distance, adaptiveness condition (treated as a factor), step (treated as a factor), NASA-TLX score, episode (treated as a factor), and number of observations. The model included a three-way interaction between mean distance, adaptiveness condition, and step, as we expected that the relationship between pupil diameter and distance varies as a function of the adaptiveness and step conditions. Random intercepts were included for each participant to account for individual differences. Using histograms and scatterplots, we confirmed the model assumptions, i.e., the normality of the distribution of residuals and similar variance (homoscedasticity). To further assess the relationship between self-reported CWL and robot adaptiveness (i.e., relationship (4)), we performed a separate paired sample t-test.

Regarding relationship (5) from Section~\ref{Sec3.1}, the next series of analyzes explored whether \emph{pupil dilation varied as a function of adaptiveness at critical time points}. Specifically, we focused on critical moments when participants were approximately 20 cm away from the robot. A custom R script processed pupil data around these key time points. The script read critical time points from a CSV file containing data for all 16 participants (M = 4.75 critical points per participant, SD = 1.39). We extracted pupil dilation data for each participant in a ±2-second window for each critical time point, sampling at 100 Hz, resulting in 400 data points per critical moment, and we performed a baseline correction to account for individual differences in baseline pupil size. We calculated the mean pupil diameter before t0 for each participant and adaptiveness condition, and subtracted this baseline value from all diameter measurements with the corrected pupil diameter reflecting changes relative to the pre-critical baseline.
To analyze the effects of adaptiveness and time on pupil dilation, we focused on the 2 seconds following the critical time point. We estimated a linear mixed-effects model using the \texttt{lme4} package in R, with corrected pupil diameter as the dependent variable. Fixed effects included adaptiveness condition (adaptive vs. non-adaptive), relative time position (1 to 200 sample points), and their interaction. To account for individual differences, we included random intercepts for participants. This model allowed us to examine how pupil dilation changed over time after the critical moment and whether this change differed between the adaptive and non-adaptive conditions while controlling for between-subject variability.

We furthermore explored the system usability corresponding to relationship (6) from Section~\ref{Sec3.1} using the data analysis tool provided by the UEQ-S\footnote{See \url{https://www.ueq-online.org/}}. In addition to the mean values per item and a paired sample t-test, we use and report their benchmark analysis for each condition (adaptive / non-adaptive) separately. 

Finally, we have asked each participant to describe the adaptive behavior in his/her own words, how they felt in each condition, and which one they prefer to assess relationship (7) in Section~\ref{Sec3.1}. These qualitative interviews followed an open questionnaire, and the data is analyzed by summarizing the substantive topics and content per question for each participant. Given the sample size, we focus on overall frequencies and highlight re-occurring relevant key statements. 

To assess relationships (1)-(7) appropriately requires consideration of the actual speed levels and speed reductions the adaptive system executed. The average speed in the adaptive condition during all interactions was 67.25\% of the constant speed level performed in the non-adaptive condition. During all interactions of the 16 participants with the adaptive system, the system ran at 100\% of the non-adaptive speed for 28.51\% of the time, at 80\% for 25.58\% of the time, at 50\% for 22.55\% of the time, and at 30\% for 23.55\% of the time. The positive effects on the users' evaluations therefore need to be interpreted in consideration of a potential 32.74\% negative impact on productivity.


\subsection{Results}\label{Results}

We report the results from our evaluation according to the relationships (1)-(7) in Section~\ref{Sec3.1}.

\paragraph{(1) Pupil Diameter and Distance to the Robot} Our linear mixed-effects model showed several significant effects on pupil diameter. First, we observed a main effect of mean distance, F(1, 159.41) = 17.13, p $<$ .001, $\eta p^{2}$ = 0.097, indicating that pupil diameter varied significantly with spatial distance, such that pupil diameter decreased as distance increased. These findings indicate, as predicted, that participants experience lower cognitive load further away from the robot, $\beta = -3.19, SE = 0.77, t = -4.14, p < .001$. 
The step (Trial 1/2 \& Assembly 1/2) also revealed a significant main effect, F(3, 165.83) = 16.16, p $<$ .001, $\eta p^{2}$ = 0.226, suggesting that pupil diameter changed across different stages of the task. Notably, pupil diameter was smaller in the second half of the task (Assembly 1 = 34.82, Assembly 2 = 34.85) relative to the first half (Trial 1 = 36.96, Trial 2 = 37.04). This pattern suggests a potential decrease in cognitive load as participants became more familiar with the task, indicating increasing familiarity over the duration of the task. 

\begin{figure*}[t]
\includegraphics[width=\textwidth]{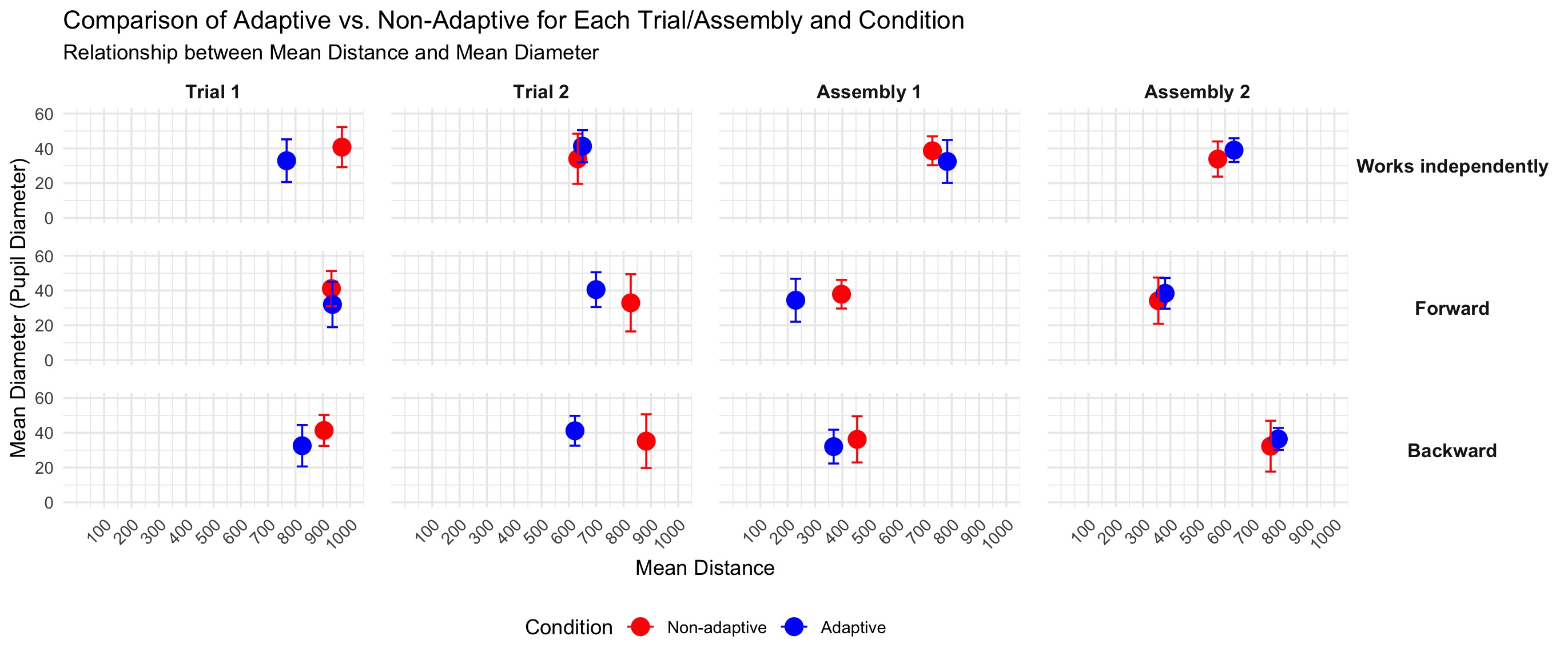}
    \caption{Mean Distance and Pupil Diameter of Operators Across the 4 Steps and 3 Episodes of the Evaluation.}
    \Description{The figure depicts average pupil diameter as a function of average distance across the 4 steps (Trial 1/2 \& Assembly 1/2 and 3 episodes (Works independently, Forward, Backward). Blue dots represent the adaptive condition, and red dots represent the non-adaptive condition. Error bars depict the 25th and 75th percentiles.}
    \label{Results1}
\end{figure*}

\paragraph{(2) Pupil Diameter and Robot Adaptiveness}
While the adaptiveness condition did not show a significant main effect, F(1, 173.54) = 0.63, p = .428, $\eta p^{2}$ = 0.003, it produced a set of significant and expected interactions: First, we observed a significant two-way interaction between mean distance and adaptiveness, F(1, 159.39) = 14.19, p $<$ .001, $\eta p^{2}$ =0.081, as well as between adaptiveness and step, F(3, 166.03) = 10.08, p $<$ .001, $\eta p^{2}$ = 0.154, and finally, a significant three-way interaction between mean distance, adaptiveness, and step, F(3, 159.39) = 6.01, p $<$ .001, $\eta p^{2}$ =0.101. To explore the nature of the three-way interaction between mean distance, adaptiveness, and step, we conducted separate linear mixed-effects models for each step (Trial 1/2 \& Assembly 1/2) of the task. The analysis indeed confirmed our earlier assertion that the interaction between mean distance and adaptiveness reached statistical significance only in 'Trial 1', F(1, 31.74) = 6.42, p = .016, $\eta p^{2}$ =0.168. For all other steps (Trial 2, Assembly 1, Assembly 2), the interaction between mean distance and adaptiveness did not reach statistical significance (all p’s $>$ .2). These findings highlight that the role of adaptiveness is especially important in the early onset of the interaction with the robotic assembly arm. Finally, we examined the main effect of distance separately for the adaptive and non-adaptive conditions in Step `Trial 1'.
The analysis revealed no significant effect of distance on pupil diameter in the adaptive condition, F(1, 15.17) = 1.16, p = .299, $\eta p^{2}$ = 0.07, but this effect was significant in the non-adaptive condition, F(1, 15.99) = 6.41, p = .022, $\eta p^{2}$ =0.286, where the mean pupil diameter was significantly smaller with increased distance from the robot, $\beta$ = -2.12, SE = 0.82, t = -2.56, p = .021 (see Figure \ref{Results1}). In other words, while the regression analysis showed that pupil diameter varied with distance in the non-adaptive condition, it was stable across distances in the adaptive condition. In addition, the mean pupil diameter was smaller overall in the adaptive condition (M = 33.99, SD = 9.37) compared to the non-adaptive condition (M = 39.93, SD = 7.01).

\paragraph{(3) Self-reported CWL and Pupil Diameter}

The NASA-TLX score did not show a significant main effect above the significant experimental factors, F(1, 170.78) = 0.51, p = .475, $\eta p^{2}$ = 0.002, suggesting that subjective cognitive load alone did not significantly predict pupil diameter.

\paragraph{(4) Self-reported CWL and Robot Adaptiveness}

A paired-sample t-test revealed a significant difference of the NASA-TLX scores in the adaptive (M = 8.36, SD = 2.94) and non-adaptive (M = 9.65, SD = 3.65) conditions, t(15) = 2.81, p = .013, Cohen’s d = 0.70. The mean difference between conditions was sizable, 1.29, 95\% CI [0.31, 2.27], with the adaptive condition showing significantly lower NASA-TLX scores. These results suggest that the adaptive condition was associated with a significantly lower perceived cognitive load than the non-adaptive condition.

\paragraph{(5) Pupil Diameter and Critical, Close-Proximity Situations}

Our analysis of critical interaction situations as introduced above reveals a significant main effect of the adaptiveness condition, F(1, 5839.26) = 17.22, p $<$ .001, $\eta p^{2}$ =0.002, indicating that pupil dilation was significantly smaller for the adaptive condition (baseline corrected score across all 200 sample points M = -0.835) relative to the non-adaptive condition (M = -0.361). There was also a significant main effect of relative time position, F(1, 5834.10) = 20.42, p $<$ .001, $\eta p^{2}$ =0.003, demonstrating that the pupil diameter decreased further away in time from the critical time point, $\beta$ = 0.01, SE = 0.001, t = -4.51, p $<$ 0.001.  Importantly, a significant interaction between adaptiveness condition and relative time position was observed, F(1, 5834.10) = 4.63, p = .031, $\eta p^{2}$ =0.001.  
We conducted separate linear mixed-effects models for each adaptiveness condition to further investigate the significant interaction between the adaptiveness condition and relative time position. In the adaptive condition, we found no significant effect of relative time position on corrected pupil diameter, F(1, 3011.25) = 2.19, p = .139, $\eta p^{2}$ =0.001. Conversely, in the non-adaptive condition, we observed a significant effect of relative time position on corrected pupil diameter, F(1, 2809.09) = 28.66, p $<$ .001, $\eta p^{2}$ = 0.01. Pupil diameter reduced significantly further away from the critical time point, $\beta$ = -0.01, SE = 0.0009, t = -5.35, p $<$ 0.001. Similar to the effects found in ‘Task 1’ above, these results confirm our previous results, after which the pupil diameter remained smaller and more stable in the adaptive condition.

\paragraph{(6) Usability and Robot Adaptiveness} 

\begin{figure*}[t]
    \centering
    \includegraphics[width=0.7\linewidth]{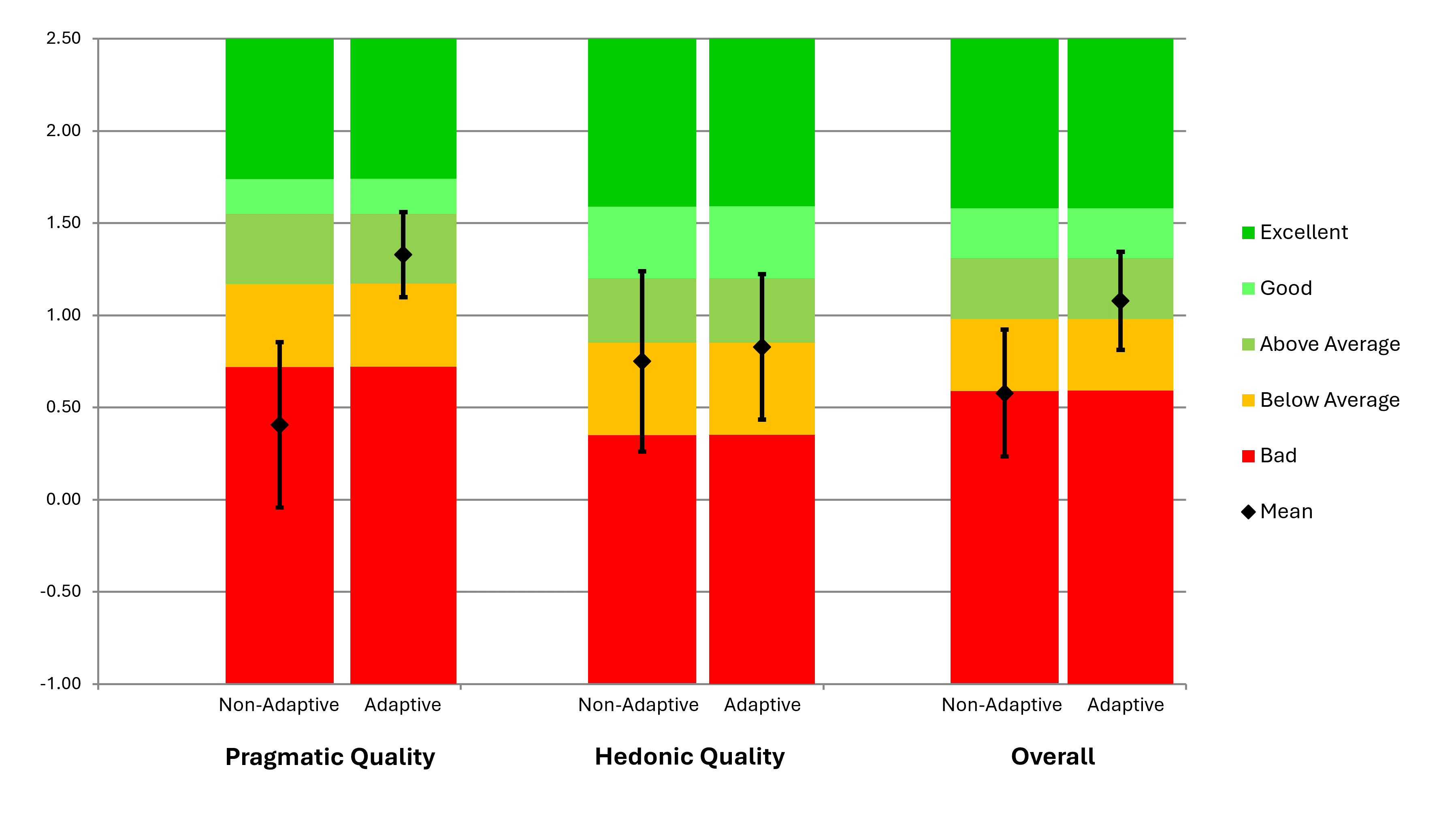}
    \caption{UEQ Benchmark Analysis for the Adaptive and Non-Adaptive Conditions of the Evaluated System.}
    \label{Benchmark}
    \Description{The figure depicts bar charts comparing pragmatic and hedonic quality as well as the overall usability, and shows a benchmark to other systems that have used the UEQ. For pragmatic quality, the adaptive system was rated above average, whereas the non-adaptive system was rated bad. For hedonic quality, both systems were rated below average, and overall, the adaptive system is rated above average whereas the non-adaptive system is rated below average to bad.}
\end{figure*}

The UEQ-S data analysis tool separates user experience in hedonic and pragmatic qualities. Hedonic usability includes non-task-oriented aspects such as originality and innovativeness, whereas pragmatic quality represents a more task-oriented usability. Values between -0.8 and 0.8 represent a neutral evaluation, and values $>$0.8 ($<$-0.8) represent positive (negative) evaluations (see Table~\ref{tab:UEQ}). In the non-adaptive condition, participants rate the system's pragmatic quality with 0.406 and hedonic quality with 0.750, with an overall score of 0.578 (see Table~\ref{tab:UEQ}). In the adaptive condition, participants rated the system's pragmatic quality with 1.328 and hedonic quality with 0.828, with an overall score of 1.078 that was significantly higher than for the non-adaptive condition. Paired-samples t-tests were conducted to compare the pragmatic and hedonic quality ratings, respectively, between the adaptive and non-adaptive groups. For the pragmatic ratings, there was a significant difference in the scores for the adaptive (M = 1.33, SD = 0.47) and non-adaptive (M = 0.41, SD = 0.92) conditions; t(15) = 3.17, p = .006. The mean difference was 0.92 (95\% CI [0.30, 1.54]). Cohen's d was 0.79 (95\% CI [0.22, 1.35]), indicating a medium to large effect size. For the hedonic ratings, there was no significant difference in the scores for the adaptive (M = 0.83, SD = 0.80) and non-adaptive (M = 0.75, SD = 1.00) conditions; t(15) = 0.54, p = .600. The mean difference was 0.08 (95\% CI [-0.23, 0.39]). Cohen's d was 0.13 (95\% CI [-0.36, 0.62]), indicating a small effect size. These findings reveal that participants evaluated the pragmatic quality as significantly higher in the adaptive compared to the non-adaptive experimental condition, highlighting that the subjective value stems especially from the perception of the adaptive system as more supportive, easy, efficient and clear. This finding is also important as it directly rules out that the current findings can be merely explained by a `novelty effect' such that participants might have perceived the adaptive system simply as more exciting or novel (which would rather impact the hedonic quality score). Figure~\ref{Benchmark} compares the two conditions to a data set from 21175 persons from 468 studies, provided by the UEQ analysis tool\footnote{\url{https://www.ueq-online.org/Material/Short_UEQ_Data_Analysis_Tool.xlsx}}. This highlights, in particular, the comparatively superior pragmatic quality of the adaptive system.

\begin{table*}[]
 \caption{UEQ-S Items per Condition}
    \label{tab:UEQ}
    \Description{The table contains the ratings of the 8 UEQ-S items for the adaptive and the non-adaptive condition. In addition, it lists the negative/positive word pairs, and assigns these to pragmatic/hedonic quality.}
\centering
\resizebox{\textwidth}{!}{%
\begin{tabular}{l|lll|ccc|lll}
              & \multicolumn{3}{c|}{\textbf{Adaptive Condition}}       & \multicolumn{3}{c|}{\textbf{Non-adaptive Condition}}                                                                &                   &                   &                   \\
\textbf{UEQ-S Item} & \textbf{Mean} & \textbf{Variance} & \textbf{Std. Dev.} & \multicolumn{1}{l}{\textbf{Mean}} & \multicolumn{1}{l}{\textbf{Variance}} & \multicolumn{1}{l|}{\textbf{Std. Dev.}} & \textbf{Negative} & \textbf{Positive} & \textbf{Scale}    \\ \hline
1             & 0.6           & 1.2               & 1.1                & -0.7                              & 1.3                                   & 1.1                                     & obstructive       & supportive        & Pragmatic \\ \hline
2             & 1.6           & 1.3               & 1.1                & 0.4                               & 1.6                                   & 1.3                                     & complicated       & easy              & Pragmatic \\ \hline
3             & 0.9           & 1.0               & 1.0                & 0.8                               & 1.4                                   & 1.2                                     & inefficient       & efficient         & Pragmatic \\ \hline
4             & 2.1           & 0.8               & 0.9                & 1.2                               & 2.3                                   & 1.5                                     & confusing         & clear             & Pragmatic \\ \hline
5             & 0.7           & 1.6               & 1.3                & 0.9                               & 2.1                                   & 1.4                                     & boring            & exciting          & Hedonic   \\ \hline
6             & 0.8           & 1.1               & 1.0                & 0.9                               & 1.6                                   & 1.3                                     & not interesting   & interesting       & Hedonic   \\ \hline
7             & 1.0           & 1.1               & 1.0                & 0.6                               & 2.0                                   & 1.4                                     & conventional      & inventive         & Hedonic   \\ \hline
8             & 0.8           & 1.2               & 1.1                & 0.6                               & 1.2                                   & 1.1                                     & usual             & leading edge      & Hedonic   \\ \hline
\end{tabular}%
}
\end{table*}

\paragraph{(7) Qualitative Perception and Robot Adaptiveness}

For our qualitative analysis, we first asked participants about the observed differences between the conditions, detached from how they perceived them. As all participants were introduced to the movement adaptations, we asked them to focus on their perception while they were performing the assembly task. Nine participants explicitly mentioned that they have noticed speed differences between the conditions and/or speed reductions in the adaptive condition. Six participants additionally noticed trajectory changes. As the Lego task was quite engaging, five participants mentioned that they have not consciously noticed any adaptive behavior. We then asked participants about how they felt in each condition after each task was completed. These qualitative interviews revealed the following key insights:

\begin{itemize}
    \item \emph{Feeling of Comfort / Safety / Trust / Stress Reduction:} 15 participants mentioned that the adaptive conditions made them feel more \textit{comfortable} than the non-adaptive conditions. 6 participants additionally mentioned that they felt more \textit{safe} in the adaptive condition, and 5 participants additionally mentioned explicitly that they were more \textit{stressed} in the non-adaptive condition, while they reported that the adaptive behavior effectively reduced their stress levels: \emph{P12: ``The non-adaptive robot was just doing its thing, it was fast even though I was working there.''} and \emph{P3: ``The adaptive robot felt more safe, I had no problem with keeping smaller distances.''} 3 participants mentioned that they \emph{trusted} the adaptive robot more than the non-adaptive robot. One of these three participants mentioned that they trusted it more because they appreciated the programmer's consideration of their personal safety: \emph{P9: ``I felt safer with the adaptive robot because I knew, it is programmed to make me feel safer. It shows that the programmer considers my safety.''}

    \item \emph{Threat / Collision Avoidance:} One explanation for the feeling of comfort and safety in the adaptive condition was that participants were not afraid of \textit{collisions}. In the non-adaptive condition, 4 participants mentioned that they had to get around the robot which was stressful and threatening: \emph{P14: ``I felt more stressed in the non-adaptive condition, I had to make sure the robot doesn’t attack me.''} and \emph{P3: ``The non-adaptive robot ignored me and moved fast, I was scared that it hits me.''}

    \item \emph{Perceived Collaboration:} 3 participants mentioned that working next to the adaptive robot made them feel working \textit{collaboratively}. In contrast, working with the non-adaptive robot made one participant feel to just do the robot's groundwork: \emph{P12: ``It felt more supportive in the adaptive condition, in a collaborative way.''} and \emph{P8: ``With the adaptive robot it felt collaborative, with the non-adaptive more like separated.''}
    
    \item \emph{Comfort vs. Task Performance:} Three participants mentioned that they do prefer the adaptive robot, but that it depends on the expected \textit{performance}: \emph{P6: ``It depends on the performance target, the adaptive robot might be too slow.''}. Two additional participants mentioned that they prefer to trade task performance against comfort: \emph{P10: ``I felt more relaxed in the adaptive condition, but the non-adaptive robot would be more efficient.''} and \emph{P5: ``I think that reducing speed for less stress is a good trade-off.''}

    \item \emph{Robot Sound:} Since the task was quite engaging for some participants, not all of them have perceived the robot's adaptiveness while working next to it. Still, three participants have mentioned the robot's \textit{sound} which was louder and more scary in the non-adaptive condition: \emph{P11: ``The adaptive robot felt more comfortable because it was quieter. The sound of the non-adaptive robot stressed me.''} and \emph{P4: ``The non-adaptive robot was louder and sounded dangerous which distracted me from the task.''}

    \item \emph{Predictability:} Two participants mentioned predictability, but expressed opposing perceptions. One participant (P2) liked the predictability of the non-adaptive robot, given its constant speed and trajectory. Another participant however mentioned the adaptive robot to be more predictable: \emph{P5: ``The adaptive robot felt more comfortable as it was more predictable.''}

    \item \emph{Preference:} When asked about their preference, 15 out of 16 participants would prefer to work with the adaptive robot. 3 out of these 15 participants mentioned that if the performance target would be high, they would probably prefer the non-adaptive robot.
    
\end{itemize}

\section{Discussion, Implications and Recommendations} \label{Sec4}

In this work, we have defined, selected (both Section~\ref{sec2}), implemented, and evaluated (both Section~\ref{sec3}) system components---measurements, relationships, and movement adaptations---for a real-time adaptive industrial robot system. Our user study revealed the following key findings:

\begin{itemize}
   
    \item Pupil diameter demonstrates a generally negative relation to distance to the robots, indicating higher workload in close proximity to the robot and vice versa.

    \item When starting an interaction with a robot, the adaptive condition compensates the effect of close distances on pupil dilation, whereas in the non-adaptive condition, smaller distances to the robot lead to increased pupil dilation.
        
    \item Pupil data reveals a familiarity effect such that after 5-10 minutes of interaction with our system, participants display decreasing pupil diameters (and hence possibly lower workload) in both conditions.

    \item After critical, close distance situations, pupil diameter remained smaller and more stable in the adaptive condition, compared to larger pupil diameter and more variation with distance to the robot in the non-adaptive condition.
        
    \item Subjective questionnaire data (NASA-TLX and UEQ-S) show significantly lower workload and higher usability in the adaptive condition.

    \item In the qualitative assessment, the vast majority of participants prefer the adaptive conditions. Positive qualitative evaluations suggest higher feelings of comfort, safety, trust, and a heightened sense of collaboration in the adaptive condition. The non-adaptive robot was evaluated as more efficient while predictability was perceived inconsistently.
\end{itemize}

\subsection{Relevant Factors Triggering System Adaptations}
\label{InputDiscussion}
 
We have discussed and collected several objective and subjective human responses during the interactions with our adaptive robot, and analyzed their relations to users' distances to the robot and the robot's adaptiveness. Our approach provides three potential developments that enrich existing study designs in HRI with regard to the consideration of user data: 

\begin{itemize}
    \item A content-wise shift from static to dynamic user characteristics,
    \item a collection-wise modification from point-in-time to continuous data collection, and
    \item an objectification by focusing on objective rather than subjective, self-reported human responses.
\end{itemize}

Existing findings in HRI on individual differences mostly focus on static characteristics such as age, gender or prior experience with robots. As findings on these individual differences are inconsistent and cannot inform a dynamic system at run time, we have demonstrated and propose to further investigate behavioral and physiological measures to differentiate users, and to effectively trigger system adaptations. Specifically, our adaptation to spatial differences have led to improved evaluations of the robot, and our analysis of pupil diameter patterns reveals the suitability of pupillometry to detect individual stress-related human responses during such interactions. 

Regarding real-time adaptiveness, we require more temporal data variance, and our study demonstrates the relevance of dynamic user characteristics that can be observed and collected at run time. We have implemented adaptive behaviors to the distance between the robot and the user, and we have simultaneously collected users' pupil data. As these measures vary during interactions, they are not only collectable at run time, but also disclose the human's states continuously.  

Regarding our requirement to non-intrusively collect dynamic human responses and to focus on the related objectiveness of these measures, we have compared subjective and objective evaluations of interactions with our prototype. Most studies in HRI use questionnaires to evaluate interactions with robots. Our comparison of pupil dilation with self-reported CWL-perceptions revealed that the NASA-TLX results alone do not allow to predict pupil diameter. However, we found promising relations between subjective measures and the collected objective data that might enable to replace self-reported questionnaires in the future, additionally preventing self-report biases. 

During the interactions with our prototype, we have found a familiarity effect that seems to affect pupil dilation during a rather short interaction of 10-20 minutes. Still, in critical, close-proximity situations, we found relevant pupil dilation patterns that further emphasize the suitability of pupillometry for future adaptive systems.

We derive the following implications and recommendations.




\vspace{1em}
\begin{mdframed}

Implications and Recommendations 
\begin{itemize}
    \item Investigate pupil dilation to directly trigger robot behavior adaptations.
    \item Evaluate additional human and context factors as well as appropriate technologies systematically (see Section~\ref{sec2}).
    \item Replicate isolated parts of our system and study the causal relationships for separate human- and context-factors in more detail.
    \item Conduct longitudinal studies to investigate the effect of repeated interactions with a robot system.
\end{itemize}
\end{mdframed}

\subsection{Robot Movements and Adaptive Behavior}\label{OutputDiscussion}

Following the presented results, we believe that developments of future robots and future HRI research will result in additional movement behaviors that can be further investigated. Taking into account psychological safety, trust, and user experience considerations will likely increase the potential of preferred outcomes by using more behavioral adaptations when developing adaptive robot systems. However, the movements themselves as well as the effect of certain movement adjustments have additional consequences that need to be considered. As described above, we include robot speed and movement range. Following the explanations in~\cite{Hostettler2022}, robot speed can be adjusted without influencing the robot's trajectory which would also be feasible with movement smoothness. Adjustments of movement range obviously change the robot's trajectory, though keeping the start and endpoint of a movement the same. Thus, with regard to the organizational goal of productivity gains, speed and movement range do affect task cycle times whereas smoothness would do so only insignificantly. The speed reduction caused by the system's adaptiveness functionality leads to a potential decrease in productivity of 32.74\%. In the qualitative assessment of our adaptive robot system, participants highlighted this trade-off. Working with the adaptive robot felt stress-reducing, however, participants mentioned that if they had to work faster, the speed reduction might hinder their performance. Being a primary goal of our system to increase ergonomics and operator satisfaction, it might however be appropriate to accept a possibly lower turnover in exchange for higher employee happiness. 

With regard to other movement-related research fields in HRI, we have not explicitly investigated the predictability of the robot movements and movement adaptations performed by our prototype, even though we found some indications in the qualitative user evaluations. This requires more research as we assume that some movement adaptions we have implemented affect each other, for example, the visibility and perception of movement adaptions might affect each other in terms of overlay, such as vanishing visibility of trajectory changes if the robot runs at full speed. In addition, sudden changes of movements might affect operators differently than gradual changes, and have an effect on relevant concepts such as predictability.

We derive the following implications and recommendations for future research on adaptation behaviors.

\vspace{1em}
\begin{mdframed}
    
Implications and Recommendations
\begin{itemize}
    \item Implement additional movement behaviors that potentially influence preferred outcomes for humans.
    \item Investigate effects of simultaneous movement adaptations (e.g. speed and trajectory changes) and adaptation strength in more detail.
    \item Evaluate trade-offs between task performance of the robot and preferred behaviors by humans, comparing individual and organizational points of view.
\end{itemize}
\end{mdframed}

\subsection{Implementation Challenges}\label{SystemMechanismDiscussion}

As outlined above, the variety of human- and context-factors and movement parameters comes with several interdependencies. These interrelations not only complicate the technical implementation of a real-time adaptive system, but also necessitate sophisticated analysis capabilities. Regarding the autonomy of such systems in future, including additional machine-learning capabilities such as \emph{RoboCat: A self-improving robotic agent} presented in~\cite{bousmalis2023robocat} might enable additional autonomy by independently detecting relevant real-time data and adaptation until satisfying user states are reached. Enlarging the number of human- and context-factors, and in particular collecting dynamic data requires different sensors that deliver a variety of data formats. To take this variety into account, different data streams have to be analyzed differently based on their temporal and functional characteristics. With regard to temporal challenges, relevant lead times of certain data streams need to be defined. For example, to detect relevant spikes in pupil dilation, we might analyze a few-second window, whereas static data such as the user's experience can be included within milliseconds, as it does not change during the interaction. In addition, different eye trackers' sampling rates reach from 30 up to 1'000 HZ, and device-specific lead times have to be chosen accordingly. 

To consider these implementation-related factors, we propose the following implications and recommendations.

\vspace{1em}
\begin{mdframed}

Implications and Recommendations
\begin{itemize}
    \item Develop analysis methods that evaluate pupil data at run time and are constructed as modules for integration with larger systems.
    \item Implement frameworks that allow to include and configure different types of sensors while considering their data formats and temporal stream characteristics.
    \item Develop and investigate machine learning approaches that autonomously detect relevant human and context data during interactions with technologies.
\end{itemize}
\end{mdframed}

\subsection{Limitations}
\label{Limitations}

Our system and evaluation come with several noteworthy limitations. Regarding the user study, we have tested our system with a rather small and homogeneous sample from a university environment and with limited cultural diversity. Even though technical affinity in our sample was rather high, most of the participants have not interacted with articulated robots on a regular basis, and represent novices. Future studies should replicate our results with actual robot operators in manufacturing companies to generalize the presented results. Still, we have demonstrated that our implemented adaptiveness functionality improves the users' evaluation of the system in the beginning of the interaction which represents an additional benefit for first time users, and we believe that this first impression might have large effects on overall robot acceptance.

Similarly, we have implemented and tested our system in a lab environment. Even though this comes with several advantages with regard to external influences, it limits the generalizability for real-life use cases. We have implemented a simple HRC situation with a focus on non-intrusive data collection. Even though this resembles actual shopfloor tasks and conditions, the suitability of our setup requires further investigations in the industrial manufacturing context regarding tasks, efficiency, robot types, and the used technical devices. In addition, the trade-off between employee well-being and the efficiency targets of an organization requires further investigation.

Regarding our selection of human factors and movement behaviors, we have focused on a practicable and testable set of factors, but we do not claim the lists of factors presented and evaluated to be exhaustive. We rather see our prototype as a starting point to achieve awareness and reactiveness for future intelligent systems. To generalize our approach to other interactive devices and settings, the used technologies---an xArm7 robot, the Pupil Core eye tracker, and the Intel RealSense camera---represent, again, a starting point for future systems that include more and/or other brands, devices, and technologies.


\section{Conclusion}\label{Sec5}

We presented an adaptive robot system that monitors and responds to human physiological and behavioral signals, specifically pupil dilation and proxemics behavior, to study the potential of human interaction with real-time adaptive robots. We demonstrated the effectiveness of adaptive systems empirically through a user study, showing that this system leads to a significant reduction in both, subjective and objective cognitive workload while improving usability, comfort, safety, and trust compared to a non-adaptive condition. The ability to use real-time behavioral and physiological data to drive robot behavior and analyze its effects on a human operator represents a key advancement in creating more ergonomic and psychologically safe human-robot collaborations.

Our findings highlight the potential of non-intrusive, real-time physiological sensing—particularly pupil dilation as an effective means to adjust robot movement behavior dynamically and has the potential to open up new directions in HCI more broadly. By reducing perceived levels of stress and increasing task satisfaction, such adaptive systems pave the way for more intuitive and efficient interactions in industrial settings. Moreover, our analysis methodology offers a framework for future research, providing a robust approach to evaluating adaptive systems and benchmarking ergonomic outcomes using both objective biometric data and subjective operator or user responses.

Looking ahead, our work opens several avenues for future adaptive systems with other devices and in other settings. This includes expanding the range of physiological and contextual data to inform and effectively adjust a system's behavior, investigations of the long-term effects of such systems on users, and transfers to other HCI contexts: on the road, driving assistance systems might adapt their acceleration behavior, interior lightning, seating positions and various other configurations to the individual at run-time to improve the driver’s or passengers' interaction and experience with the car; in buildings, environmental controls (e.g., lighting, heating, ventilation, air conditioning and quality control) and building functions (e.g., security systems, shading, transport systems) could provide run-time user-adapted assistance levels, thereby improving user experience and acceptance; and also software systems without physical embodiment might adapt their structure, response speeds, or interaction modalities to users' instantaneous estimated preference. We hope the evidence provided in this work will stimulate the investigation and development of these, and more, next-generation adaptive systems that create environments where humans and technologies work together seamlessly, enhancing both productivity and user well-being.

\section*{Acknowledgments}

In accordance to the ACM policy, generative AI was \textbf{not} used for creation or editing of this text.

\noindent The authors thank Artyom Zinchenko at LMU for his advice and support on data processing and statistical analysis.



\subsection*{Ethics Approval}

The Ethics Committee of the University of St. Gallen has confirmed that no ethical approval was required for this work.

\bibliographystyle{ACM-Reference-Format}
\bibliography{software}


\end{document}